# Non-destructive three-dimensional measurement of hand vein based on self-supervised network

Xiaoyu Chen[1], Qixin Wang[1], Jinzhou Ge, Yi Zhang and Jing Han*

*Nanjing University of Science and Technology, Nanjing, China*

**ABSTRACT**

At present, supervised stereo methods based on deep neural network have achieved impressive results. However, in some scenarios, accurate three-dimensional labels are inaccessible for supervised training. In this paper, a self-supervised network is proposed for binocular disparity matching (SDMNet), which computes dense disparity maps from stereo image pairs without disparity labels: In the self-supervised training, we match the stereo images densely to approximate the disparity maps and use them to warp the left and right images to estimate the right and left images; we build the loss function between estimated images and original images for self-supervised training, which adopts perceptual loss to help improve the quality of disparity maps in both detail and structure. Then, we use SDMNet to obtain disparities of hand vein. SDMNet has achieved excellent results on KITTI 2012, KITTI 2015, simulated vein dataset and real vein dataset, outperforming many state-of-the-art supervised matching methods.

## 1. Introduction

Calculating dense disparity map from stereo image pairs has always been a fundamental and classical problem in computer vision. The binocular stereo task often consists of four steps: image acquisition, epipolar rectification, stereo matching and distance calculation. The main purpose of stereo matching is to locate the same points from left and right images captured by two binocular cameras. This disparity is computed from the matched point pairs and with the help of epipolar rectification, only the horizontal matching need to be considered.

Traditional matching algorithms[20, 8, 12] are based on feature matching and do not need depth label data. Generally, these methods can be divided into four steps: matching cost calculation, cost aggregation, disparity calculation and disparity refinement. As the dense points matching in these methods are based on manual feature and simple metric strategy, the performance is limited.

In recent years, with the advance of computing power and big data, convolutional neural networks(CNNs) have provided a variety of solutions for different tasks, CNNs based on supervised learning has achieved great success in many visual tasks. In supervised CNNs for stereo matching, feature extraction, cost volume calculation and disparity estimation are integrated in an end-to-end network and disparity labels are needed for supervised training. With the advantages of unsupervised feature learning of neural networks, networks can make full use of the information provided by stereo images and thus achieve excellent results. As supervised CNNs is actually a direct estimating process, even for occlusion areas in stereo images, CNNs can get relatively accurate results. Most of the deep learning methods rely on large amounts of labeling data for training. However, obtaining disparities is a very challenging problem in some specific tasks. In non-destructive vein stereo measurement, three-dimensional labeling data is inaccessible, which makes it difficult for supervised learning.

Many methods [5, 34] attempt to use self-supervised monocular and binocular estimation to estimate disparities. Such methods eliminate the dependence on labeling disparity data and reconstruct left and right images instead of calculating disparity directly. Specifically, the networks extract left and right image features, and calculates left and right disparity firstly. Secondly, stereo images are reconstructed with the help of estimated disparity maps, and lastly, the network is trained by making reconstructed images close to reference images. From this point, this process is to find matched point pairs and similar to traditional methods. It is impossible to estimate the disparity of occlusion area in stereo images where the points cannot be matched. Besides, more complex structure makes it hard for training which leads to a big gap with supervised methods.

In summary, passive three-dimensional measurements have been widely used in indoor and outdoor clear imaging. However, supervised methods are difficult to apply to non-invasive detection of biological tissues. The main cause is that it is difficult to get accurate labels without intrusion, so self-supervised method is suitable for non-intrusive scenarios. There are three problems in non-invasive measurement of biological hand vein: firstly, absorption and scattering characteristics of biological tissues lead to low contrast of conventional imaging; secondly, three-dimensional information of hand vein is disturbed by the texture information of skin surface details, leading to poor performance of hand vein; thirdly, unsupervised methods are less accurate than supervised methods in detail and structure.

In this paper, we combine traditional matching strategy with strong feature extraction ability of deep network for self-supervised learning to learn on binocular vein images.

---

The perceptual loss function and texture removal operation are added to reduce the interference of noise on vein data. As a general matching method, our network SDMNet can adjust itself by fine-tuning and adapt to different scenes for disparity estimation.

We have three key ideas to achieve this: 1.We designed a self-supervised network for measuring hand vein that can be trained end-to-end. 2.Perceptual loss is added to loss function, occlusion areas are cropped to improve disparity optimization, and data preprocessing reduces the interference of noise and skin reflection in network training. 3.We propose a high-precision and non-destructive three-dimensional measurement scheme for vessel puncture, vascular varices and other pathological diagnosis.

## 2. Related work

Calculating dense disparity maps from rectified stereo images is a classical computer vision problem. In this section, we review the related stereo matching techniques.

Classical Stereo Matching: Traditional matching algorithms can be classified as global matching and local matching according to different methods of optimization theory. The local matching methods [26, 33] perform higher efficiency which use energy minimization strategy. However, in the energy function, there is no smoothing optimization so that areas with complex occlusion, texture and depth discontinuity are hard to estimate. On the contrary, results of global matching algorithm are more accurate, but the running time is longer which is hard to meet the requirement of real-time estimation. Generally, global matching algorithms such as belief propagation [32], graph cuts [13] compute the energy function globally and obtain optimal solution of all disparity values with leverage optimization. In order to combine the advantages of both global and local matching, Semi-Global Matching [8] adopts the global matching framework, but when calculating the of energy function minimization, the two-dimensional minimization algorithm is transformed into efficient one-dimensional path aggregation in the global framework. Semi-Global Matching method achieves great improvement in efficient in the cost of little decrease of accuracy. Post-processing methods are also used to polish the disparity results. Yang [31] checks the consistency of left and right images to enhance the robustness. MRF [16, 23] and CRF [15, 25] are also widely used in the post-processing after the emergence of tagged data sets, which improves the accuracy.

CNN-based Stereo Matching: Convolutional neural networks have achieved great success in computer vision and Mayer et al. [19] constructs an end-to-end network for estimating optical flow and disparity. The network uses an encoder-decoder structure to extract features with semantics and details and estimates disparity by regression from coarse to fine. Luo et al. [17] first replaces the regressor by classification predictor for disparity estimation. GC-Net [11] combines feature extraction, cost convolution and regularization loss function in an end-to-end network, and uses 3-D convolution to regress dense sub-pixel disparity. CRL [24] proposes a framework composed of two-stage hourglass convolution neural network to fine-tune disparity using residual network. The second-stage convolution network is used to refine the disparity map generated by the first network based on residual learning. Stereo Matching by Unsupervised Learning: In unsupervised stereo matching network, disparity estimation is transformed into image reconstruction during training with Spatial Transform Networks [9], where disparity labels are unnecessary. In Monodepth [5, 3], Disparity is generated by left-right reconstruction loss, and left-right consistency is used to constrain each other. Another way is to train an end-to-end depth estimation network of monocular images by using geometric information from different perspectives of continuous video frames [18]. Recently, according to the symmetry of left and right images, SsSMnet [36] designed a symmetric neural network, which improved the left and right consistency function and achieved the state of art result.

In summary, existing supervised methods can achieve high accuracy, but they are difficult to apply to non-intrusive detection in organisms. Unsupervised and traditional methods are suitable for scenes where ground truth is difficult to obtain, but the accuracy is hard to meet the demand in applications. From this point, we design an improved self-supervised method to improve the matching accuracy by refining details and contexts.

## 3. SDMNet

In this section, we present our self-supervised network for stereo matching, mixed loss function, improved region learning strategy. Our goal is to get dense disparity maps by matching pixels from left and right images, and refine per-pixel disparity by using perceptual loss function. The framework is shown in Fig. 1.

### 3.1. Self-supervised Framework

Our task is to predict dense disparity maps from rectified stereo image pairs. Most existing methods regard it as a supervised task, but in many scenarios it is hard to obtain dense disparity labels, such as hand vein measurement. SDMNet learns per-pixel disparity mapping from left to right: $d_l = func(I_l, I_r)$, and from right to left: $d_r = func(I_r, I_l)$. Then the left disparity is estimated and left image can be reconstructed with left disparity map: $I_l' = wrap(I_r, d_l)$. The right image can be reconstructed in the similar way: $I_r' = wrap(I_l, d_r)$. By comparing the reference image $I$, and reconstructed image $I'$, we can construct loss function to train the network. After the training process, the network can output fine disparities $(d_l, d_r)$ which can reconstruct $(I_r, I_l)$ to $(I_r', I_l')$. With estimated disparity maps, intrinsic baseline distance $b$ and camera focal length $f$, the depth can be calculated by $D_{depth} = bf/d$.

#### 3.1.1. 2D Feature Extraction

Instead of directly matching all of left and right pixels, we use deep features to improve the robustness of 1D photo-metric information matching. In our network, we use 2D convolution layers to extract the deep features of left and right images. Inspired by the very recent PSMNet [1], we use several cascaded 3 × 3 convolution kernels to expand the receptive field while reducing the amount of network parameters. Then the multi-scale information of context is extracted by feature pyramid structure (SPP) [7]. The number of the output channels in feature layer is 32, and the weights of left and right branches are shared, which satisfies the symmetry of left and right features.

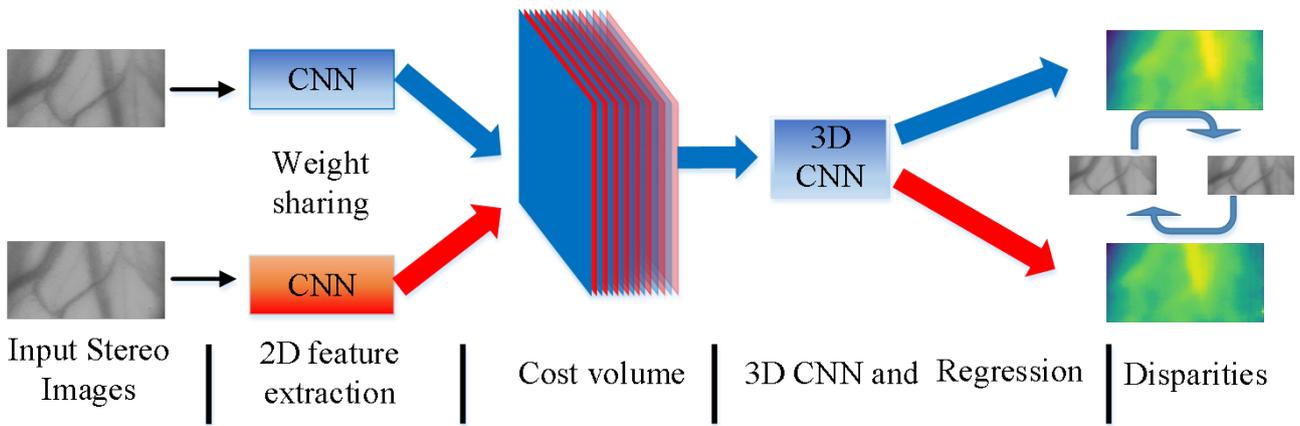

Figure 1: Our self-supervised SDMNet architecture.

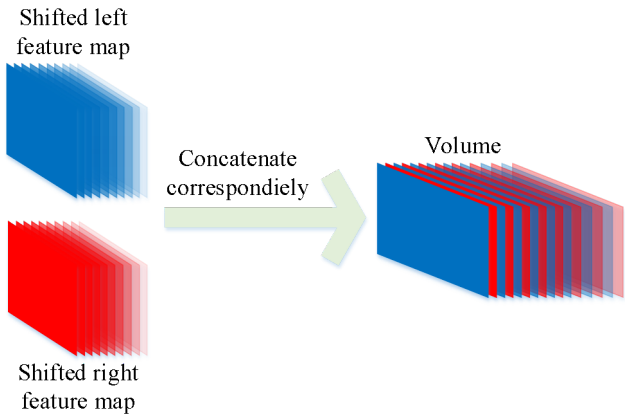

Figure 2: Feature cost volume.

### 3.1.2. Cost Volume

In order to calculate stereo matching, we use the extracted deep features to stack them into cost volumes in a preset disparity range. Unlike the two cost volumes in GC-net [11], assuming the disparity range is $D$, we form a $height \times width \times (D + 1) \times featuresize$ cost volume. As shown in Fig. 2, the blue cube represents the amputation of the left pixel map from 0 to $D$, and the red cube represents the amputation of the right pixel map from 0 to $D$. The left and right features are crossed $D + 1$ times respectively, and the left and right features are concatenated to form a final cost volume. We find that context information can be learned on this cost volume, and it works better on one-dimensional distance problem. The advantage of this structure lies in that the horizontal amputation of left and right features conforms to the physical model rules of finding matching points horizontally. At the same time, in order to reduce the computational complexity of 3D convolution, we only stack shifted features so that redundant information is removed while the information of left and right image contexts is retained. In a sense, we get the disparity maps by matching the left and right images in the horizontal dimension, and get the scene depth indirectly, instead of directly getting the depth through the features.

3D CNN and Regression: Given the cost volume, we can use the cost volume to estimate the disparity. In order to combine the 4D cost volume for estimation, height, width, inspection range and feature size, we use 3D convolution instead of 2D convolution.

### 3.2. Loss Function

Our fully self-supervised network adjusts the direction of network learning through loss function, that is, by reconstructing error constraints. Our overall loss is a combination of four parts. They are self-supervised loss, smoothness regularization loss, left-right disparity consistency loss and perceptual Loss, respectively.

$$L = w_a(l_a^l + l_a^r) + w_s(l_s^l + l_s^r) + w_c(l_c^l + l_c^r) + w_p(l_p^l + l_p^r), \quad (1)$$

among them, $l_a^l$, $l_a^r$ are reconstructed losses used to make the reconstructed image similar in pixels and structure, $l_s^l, l_s^r$ are image smoothness loss, $l_c^l, l_c^r$ are left-right consistency constraint, $l_p^l, l_p^r$ are perceptual loss. $w_a, w_c, w_s$ and $w_p$ are the weights of $l_a, l_c, l_s$ and $l_p$, respectively.

$Self-supervised\ loss$: The accuracy of disparity maps reflects indirectly on the reconstruction errors between reconstructed images and reference images. Similar to SsSMNet [36], our $l_a^l$ is defined as:

$$l_a^l = \frac{1}{N} \sum l_{a_{ij}}^l, \quad (2)$$

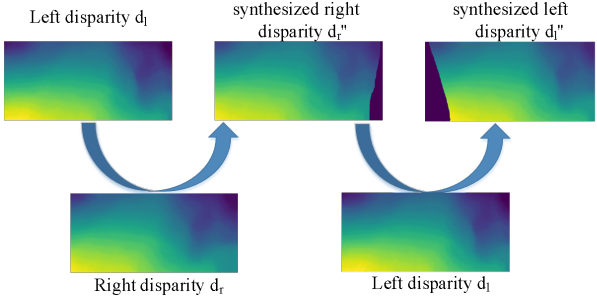

Figure 3: Loop consistency constraint structure.

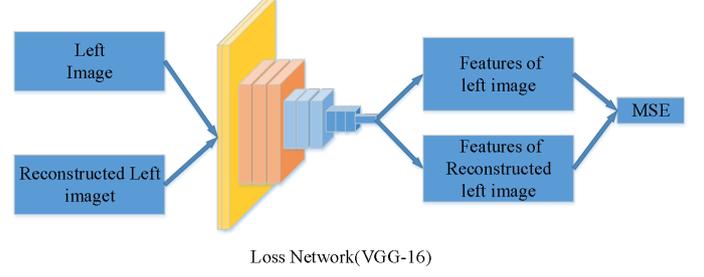

Figure 4: Refine network with Perceptual Losses.

$$l^l_{a_{ij}} = \partial_1 \frac{1 - SSIM(I^l_{ij}, I^{l\,\prime}_{ij})}{2} + \partial_2 \left| I^l_{ij} - I^{l\,\prime}_{ij} \right| + \partial_3 \left| \nabla I^l_{ij} - \nabla I^{l\,\prime}_{ij} \right|, \quad (3)$$

where $N$ is the total number of pixels in the input image, $\nabla I$ denotes the first order gradient of an image, $\partial_1$, $\partial_2$ and $\partial_1$ balance the structural similarity, brightness similarity and gradient consistency of the image. We set $\partial_1 = 0.85$, $\partial_2 = 0.15$, and $\partial_3 = 0.15$. According to STN [9] and [5], we use interpolation to reconstruct the image $I_l$ by disparity $d_l$ and reference image $I_r$. Similarly, the right image can also be reconstructed.

***Smoothness regularization loss***: For regularization terms, we consider that disparity maps need to be smoothed locally to reduce singularities. Here we use the second gradient function of the image, $l_s$ is defined as:

$$l_s = \frac{1}{N} \sum \left| \nabla^2_x d^l_{ij} \right| e^{-\left|\nabla_x I^l_{ij}\right|} + \left| \nabla^2_y d^l_{ij} \right| e^{-\left|\nabla_y I^l_{ij}\right|}, \quad (4)$$

where $\nabla^2_x$ and $\nabla^2_x$ are second-order gradient of image in horizontal and vertical directions. $e^{-\left|\nabla_x I^l_{ij}\right|}$, $e^{-\left|\nabla_x I^l_{ij}\right|}$ are smoothing factors.

### 3.2.1. Left-right disparity consistency loss

In addition to the above loss function in our model, left-right consistency loss is applied. Unlike SsSMNet [36], we use a more intuitive cycle of left-right consistency between disparity maps shown in Fig. 3. Our task is to output disparity and it is more direct to make left-right consistency of disparity.

$$l^l_c = \frac{1}{N} \sum \left| d^{\prime\prime}_{ij} - d_{ij} \right|, \quad (5)$$

the reconstructed right disparity $d'_r$ can be obtained by warping the left disparity $d_l$ to the right disparity $d_r$. The second reconstructed left disparity $d''_l$ is generated by warping $d'_r$ to $d_l$. So our consistency constraints $l^l_c$ is: With this left-right consistency cyclic loss, we can couple left and right to form a symmetrical network.

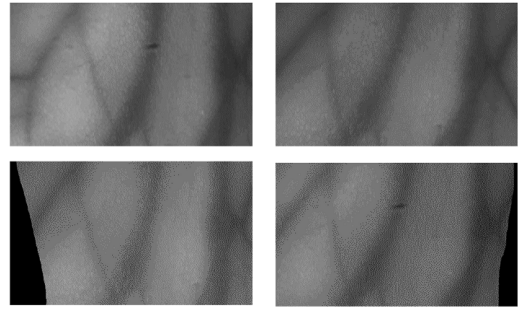

Figure 5: Region strategy. From top to bottom: left image and right image, reconstructed left and reconstructed right image.

### 3.2.2. Perceptual Loss

In pixel-level tasks, a very important idea is to extract advanced features with convolutional neural networks as an extra loss function. Inspired by Johnson et al.[10], our intuition is to iteratively refine the reconstructed image and make it similar to the reference image in structure and detail. As shown in Fig. 4, we add an extra perceptual loss [10] which calculates the semantic similarity of high level features by Mean Square Error(MSE). The features are extracted with VGG16 network [28] (extraction by vgg16 in *conv*5_3 layer) pre-trained on ImageNet1k [2]. The parameters of pre-trained model are fixed.

Thus our $l^l_p$ loss is define as:

$$l^l_p = \frac{1}{N} \sum \left\| f^l - f^{l\,\prime} \right\|^2_2, \quad (6)$$

where $f^l$, $f^{l\,\prime}$ are the left reference image and the reconstructed image obtained by VGG-16.

### 3.3. Region strategy at training time

We use STN [9] to warp horizontally, and use left-right disparity maps and input stereo images to get the reconstructed images. As shown in Fig. 5, the reconstructed images consistently show no information in the margins (black areas). If all pixels are trained, the performance of the self-supervised network will be greatly affected. Considering the disadvantages of STN transform in stereo matching, we select the effective regions to constrain, that is, only constrain the effective regions, so that the network focuses on the areas with vaid information. For that disparitys range from 0 to 160, we do not constrain the left 160 range of left disparity or the right 160 range of right disparity. We only constrain the effective region of the image selectively in the loss function mentioned above, which in fact reduce the noise in self-supervision.

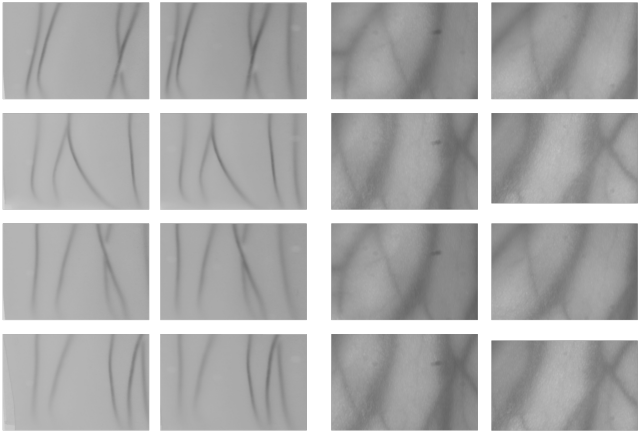

Figure 6: Simulated vein data and real vein data. In order to compare with the real blood vessel image, we show the grayscale images after the color reversal.

## 4. Data acquisition

In this section, we describe the data collection of hand vein. We collected simulated hand vein datasets (with accurate labels) and real hand vein datasets respectively. Simulated vein dataset is used for quantitative analysis and real vein datasets for qualitative analysis. In the simulated data collection, we use 3.5mm diameter lines as mimics and put acrylic plates in front of the camera to simulate skin epidermis reflection and scattering. The samples of collected datasets are shown in Fig. 6.

### 4.1. Acquisition of Stereo Image Pairs

We use two 1920*1200 CMOS monochrome cameras (BASLER's ACA 1920-155um) to shoot stereo images, and all rectified images are cropped to 256*512. The images are rectified by Zhang Zhengyou's calibration method [35]. We show part of simulated images and stereo pairs of real blood vessels in Fig. 6.

For simulated stereo images, we use 0.5mm acrylic plate to simulate the ambiguity effect of vessels. For real vein, we assembled a high pass 850nm high pass filter on a 25mm camera lens. [29] has revealed that, in 850nm band, the muscle tissue has strong reflex characteristics, strong absorption characteristics of vascular. At the same time, we use active structured light to carry out three-dimensional measurement, and the disparity label error is lower than 0.1 pixels.

## 5. Experiment

In this section, we demonstrate our experiment settings. To evaluate the performance of our method, we report the results of our vein datasets and public datasets. At the same time, we analyze perceptual loss and region strategy in self-supervised methods.

### 5.1. Experiment Details

We use four datasets in our experiment: our simulated vein dataset and real vein datasets, KITTI Stereo 2012 [4] and KITTI Stereo 2015 [21]. The proposed network is implemented with PyTorch [22]. All models were trained end-to-end with Adam($\beta_1$=0.9, $\beta_1$=0.999). All image pairs are pre-processed before entering the network. They are normalized ranging from 0 to 1. The initial learning rate of all models is 1e-3, and the learning rate after 4000 iterations is drop to 1e-4. With limited computation source, we set batch size to 1. During the training of simulated and real vein, the images are resized to = 256, = 512, and the maximum disparity $D$ range was set to 160. For KITTI, the image is randomly cut out of the original image with a size of = 256, = 512, and the maximum disparity $D$ range is set to 160. We set $w_a = 1$, $w_s = 0.1$, $w_c = 1.5$ and $w_p = 0.3$ for all experiments.

### 5.2. KITTI

Table 1
Results on KITTI 2012 stereo benchmark. Out-Noc: Percentage of erroneous pixels in non-occluded areas. Out-All: Percentage of erroneous pixels in total. Avg-Noc: Average disparity/end-point error in non-occluded areas.

| Method | Out-noc | Out-all | Avg-noc |
|---|---|---|---|
| GC-NET [11] | 1.77% | 2.30% | 0.6 px |
| Displets v2 [6] | 2.37% | 3.09% | 0.7 px |
| MC-CNN-acrt [14] | 2.43% | 3.63% | 0.7 px |
| SPS-StFl [30] | 2.83% | 3.64% | 0.8 px |
| SsSMnet [27] | 2.30% | 3.00% | 0.8 px |
| SDMNet | 2.25% | 3.00% | 0.7 px |

Our training data comes from the original image pair of KITTI, which contains 42,382 pairs of images from 61 scenes. We randomly selected 3,000 pairs to start training from scratch. KITTI-2012 consists of 194 training pairs and 195 testing pairs while KITTI-2015 contains 200 stereo pairs for training and 200 stereo pairs for testing. In Table 1 and Table 2, we evaluate the performance of our method with perceptual loss and regional constraint module on KITTI-2012 and KITTI-2015(3 pixels threshold) testing subsets respectively. Our method outperforms many state-of-the-art stereo matching methods.

In Fig. 7 and Fig. 8 we show qualitative results of our method and comparison with SsSMnet [36] on KITTI 2012 and KITTI 2015 datasets.

## Table 2
Results on KITTI 2015 stereo benchmark. D1: Percentage of stereo disparity outliers in first frame. bg: Percentage of outliers averaged only over background regions. fg: Percentage of outliers averaged only over foreground regions. all: Percentage of outliers averaged over all ground truth pixels.

| Method | D1-bg | D1-fg | D1-all |
| --- | --- | --- | --- |
| GC-NET [11] | 2.21% | 6.16% | 2.87% |
| SGM-Net [27] | 2.66% | 8.64% | 3.66% |
| CRL [24] | 2.48% | 3.59% | 2.67% |
| MC-CNN-acrt [14] | 2.89% | 8.88% | 3.89% |
| SsSMnet [36] | 2.70% | 6.92% | 3.40% |
| SDMNet | 2.74% | 6.62% | 3.39% |

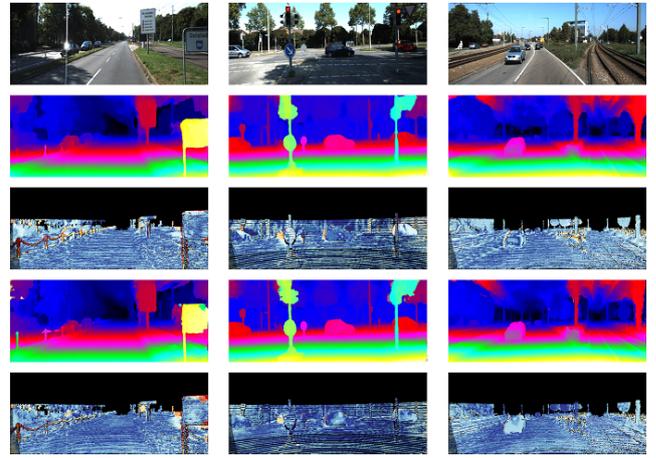

Figure 8: Our qualitative results on KITTI-2015. Top to bottom: left image, our result, our error map, result of SsSMNet [36] and its error map.

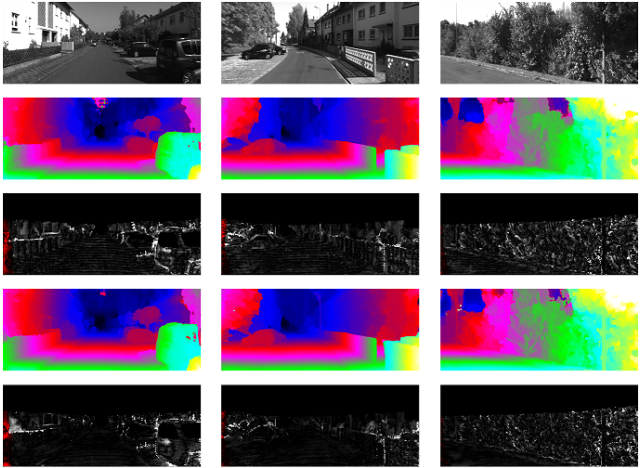

Figure 7: Our qualitative results on KITTI-2012. Top to bottom: left image, our result, our error map, result of SsSMNet [36] and its error map.

## Table 3
Comparisons of reconstruction losses.

| With P-Loss and regional constraint | Simulated vascular MAE | True vascular SSIM | L1 |
| --- | --- | --- | --- |
| Yes | 0.45 | 0.332 | 4.80 |
| No | 1.02 | 0.254 | 5.25 |

### 5.3. Test on simulated vessel datasets

We use the proposed method to carry out experiments on simulated vein dataset. We split all data for training set (40 pairs), validation (4 pairs) set and test set (4 pairs). In Fig. 9, we show qualitative testing results and error maps of our method on our simulated vessel datasets. The baseline method is implemented based on PSMNet backbone with left-right consistency loss, ssim loss and smooth loss. The baseline method can achieve a result as shown in Fig. 9 in the second row, where the foreground part in the disparity map was discontinuous and visual insignificant. Our method with perceptual loss and regional constraint can obtain a remarkably continuous disparity map.

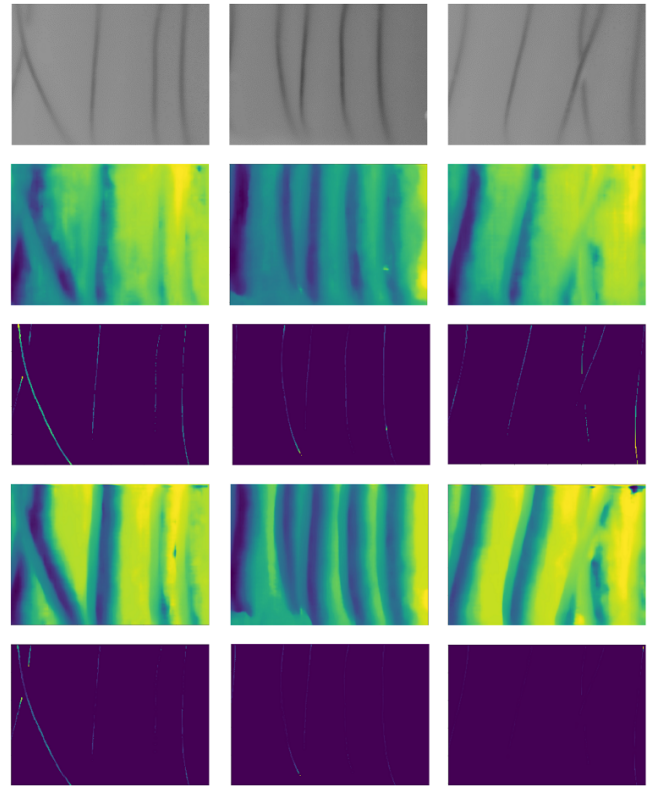

Figure 9: Qualitative evaluations on our simulated vessel data: Top to bottom: left image, result with baseline method and its error map, our result with perceptual loss and region strategy and its error map. Left gray images are visualized with color.

In Table 3, we test the performance of our model on simulated vessel ground truth data. The ground truth disparities for testing dataset are withheld for evaluation. The formula

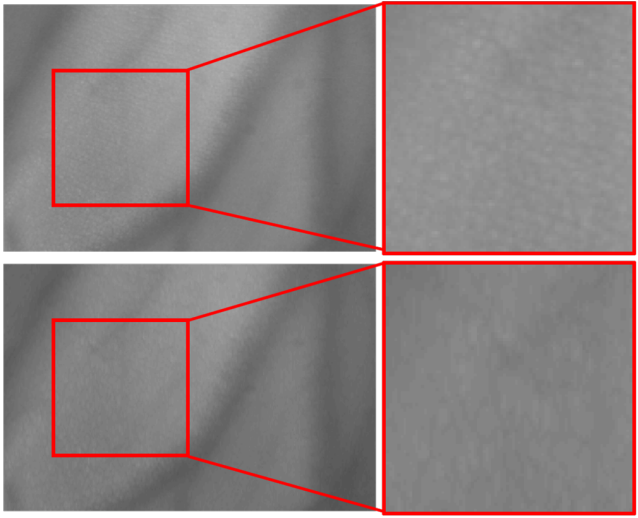

Figure 10: Gauss smoothing.From top to bottom:origin image,processed image.

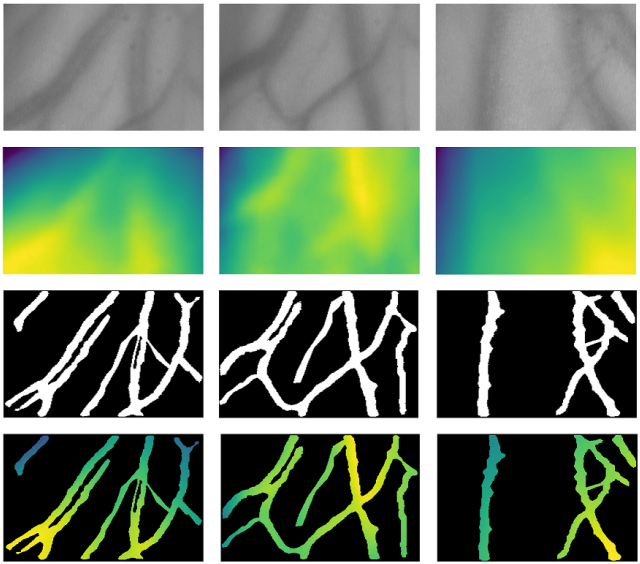

Figure 11: Qualitative on our vein data: Top to bottom: left image , left disparity , mask and disparity of vein.

MAE(Mean Absolute Deviation) is defined as:

$$MAE(X, X') = \frac{1}{N} \sum_{1}^{m} \left| X(x_{ij}) - X'(x_{ij}) \right|, \quad (7)$$

Table 3 shows that the results have been greatly improved, namely, the error metric decreases from 1.02 to 0.45 with perceptual loss and with region strategy on average. The final average error of all pixels is 0.45.

### 5.4. Test on real vessel datasets

The real collected vein dataset contains texture information of hand epidermis. In order to reduce the interference of skin texture noise on matching, we use Gauss filtering to smooth the image. The gray scale of the image is [0,255].

In order to show the effect, we display the vascular image in Fig. 10. Before processing, the epidermis texture is rich, especially at the cross-lines. It is hard to guarantee that the network matches the vein features instead of epidermis texture. After processing, the texture of the epidermis disappears, while much noise is filtered out and the image becomes more smooth.

We use the same method to train on real vessel datasets, 40 pairs for training, 4 pairs for validation and 4 pairs for testing. In Fig. 11, we show qualitative results of our method on our vein dataset. To verify the effectiveness of the proposed perceptual loss and regional constraint on real vessel datasets, we train two models with and without perceptual loss and regional strategy. With the stereo images and generated disparity maps produced by two models, we compare the similarity between reconstructed image and reference image using SSIM (higher is better) and L1 distance (lower is better) in table 3. Our method with perceptual loss and region strategy achieves higher performance on both SSIM and L1 metrics than the baseline method.

## 6. Conclusion

As the labels of hand vein are hard to obtain, we construct a self-supervised neural network for intravenous measurement which uses input stereo images for self-supervision without ground truth. A better cost volume is used to aggregate cost; a new training loss function, perceptual loss is adopted, which uses cyclic constraints in image mapping; a new region strategy is adopted which improves the convergence of the network and enhances the matching accuracy; a left-right consistency of disparity loss is added in loss function, which increase the robustness of the network. Experiments show that our method achieves high accuracy in simulated vein data. The proposed network also achieves impressive results on KITTI datasets, and even outperform some supervised methods. We will continue our research to better deal the problem of image blurring in vein data.

## Acknowledgment

The Key Research and Development programs in Jiangsu China, BE2018126.
## References

[1] Chang, J.R., Chen, Y.S., 2018. Pyramid stereo matching network, in: Proceedings of the IEEE Conference on Computer Vision and Pattern Recognition, pp. 5410–5418.
[2] Deng, J., Dong, W., Socher, R., Li, L.J., Li, K., Li, F.F., 2009. Imagenet: A large-scale hierarchical image database. Proc of IEEE Computer Vision and Pattern Recognition , 248–255.
[3] Garg, R., B.G., V.K., Carneiro, G., Reid, I., 2016. Unsupervised cnn for single view depth estimation: Geometry to the rescue, in: European Conference on Computer Vision.
[4] Geiger, A., Lenz, P., Urtasun, R., 2012. Are we ready for autonomous driving? the kitti vision benchmark suite, in: 2012 IEEE Conference on Computer Vision and Pattern Recognition, IEEE. pp. 3354–3361.
[5] Godard, C., Aodha, O.M., Brostow, G.J., 2017. Unsupervised monocular depth estimation with left-right consistency, in: Computer Vision and Pattern Recognition.

[6] Guney, F., Geiger, A., 2015. Displets: Resolving stereo ambiguities using object knowledge, in: Computer Vision and Pattern Recognition.

[7] He, K., Zhang, X., Ren, S., Sun, J., 2015. Spatial pyramid pooling in deep convolutional networks for visual recognition. IEEE Transactions on Pattern Analysis and Machine Intelligence 37, 1904–1916.

[8] Hirschmuller, H., 2007. Stereo processing by semiglobal matching and mutual information. IEEE Transactions on pattern analysis and machine intelligence 30, 328–341.

[9] Jaderberg, M., Simonyan, K., Zisserman, A., Kavukcuoglu, K., 2015. Spatial transformer networks .

[10] Johnson, J., Alahi, A., Li, F.F., 2016. Perceptual Losses for Real-Time Style Transfer and Super-Resolution.

[11] Kendall, A., Martirosyan, H., Dasgupta, S., Henry, P., Kennedy, R., Bachrach, A., Bry, A., 2017. End-to-end learning of geometry and context for deep stereo regression, in: Proceedings of the IEEE International Conference on Computer Vision, pp. 66–75.

[12] Klaus, A., Sormann, M., Karner, K., 2006. Segment-based stereo matching using belief propagation and a self-adapting dissimilarity measure, in: 18th International Conference on Pattern Recognition (ICPR'06), IEEE. pp. 15–18.

[13] Kolmogorov, V., Zabih, R., 2013. Computing visual correspondence with occlusions using graph cuts. Phd Thesis Stanford Univ 2, 508–515 vol.2.

[14] Lecun, Y., 2015. Stereo matching by training a convolutional neural network to compare image patches .

[15] Li, Y., Huttenlocher, D.P., 2008. Learning for stereo vision using the structured support vector machine, in: IEEE Conference on Computer Vision and Pattern Recognition.

[16] Li, Z., Seitz, S.M., 2007. Estimating optimal parameters for mrf stereo from a single image pair. IEEE Trans Pattern Anal Mach Intell 29, 331–342.

[17] Luo, W., Schwing, A.G., Urtasun, R., 2016. Efficient deep learning for stereo matching, in: Computer Vision and Pattern Recognition.

[18] Mathieu, M., Couprie, C., Lecun, Y., 2015. Deep multi-scale video prediction beyond mean square error .

[19] Mayer, N., Ilg, E., Hausser, P., Fischer, P., Cremers, D., Dosovitskiy, A., Brox, T., 2016. A large dataset to train convolutional networks for disparity, optical flow, and scene flow estimation, in: IEEE Conference on Computer Vision and Pattern Recognition.

[20] Mei, X., Sun, X., Zhou, M., Jiao, S., Wang, H., Zhang, X., 2011. On building an accurate stereo matching system on graphics hardware, in: 2011 IEEE International Conference on Computer Vision Workshops (ICCV Workshops), IEEE. pp. 467–474.

[21] Menze, M., Geiger, A., 2015. Object scene flow for autonomous vehicles, in: Computer Vision and Pattern Recognition.

[22] Nikolic, B., 2018. Acceleration of non-linear minimisation with pytorch .

[23] Pal, C.J., 2012. On learning conditional random fields for stereo. International Journal of Computer Vision 99, 319–337.

[24] Pang, J., Sun, W., Ren, J.S., Yang, C., Yan, Q., 2017. Cascade residual learning: A two-stage convolutional neural network for stereo matching .

[25] Scharstein, D., Pal, C., 2007. Learning conditional random fields for stereo, in: IEEE Conference on Computer Vision and Pattern Recognition.

[26] Scharstein, D., Szeliski, R., 1998. Stereo matching with nonlinear diffusion. International Journal of Computer Vision 28, 155–174.

[27] Seki, A., Pollefeys, M., 2017. Sgm-nets: Semi-global matching with neural networks, in: IEEE Conference on Computer Vision and Pattern Recognition.

[28] Simonyan, K., Zisserman, A., 2014. Very deep convolutional networks for large-scale image recognition. Computer Science .

[29] Wu Wei, Yuan Wei Qi, L.S.K.D., Hongtao, Z., 2012. Selection of typical wavelength for palmar vein recognition. Acta Optica Sinica 32, 133–139.

[30] Yamaguchi, K., Mcallester, D., Urtasun, R., 2014. Efficient joint segmentation, occlusion labeling, stereo and flow estimation, in: European Conference on Computer Vision.

[31] Yang, Q., 2015. Stereo matching using tree filtering. IEEE Transactions on Pattern Analysis and Machine Intelligence 37, 834–846.

[32] Yang, Q., Wang, L., Yang, R., StewaNius, H., NistaR, D., 2009. Stereo matching with color-weighted correlation, hierarchical belief propagation, and occlusion handling. IEEE Transactions on Pattern Analysis and Machine Intelligence 31, 492–504.

[33] Yoon, K.J., Kweon, I.S., 2006. Adaptive support-weight approach for correspondence search. IEEE Trans Pattern Anal Mach Intell 28, 650–656.

[34] Zhang, Y., Khamis, S., Rhemann, C., Valentin, J., Kowdle, A., Tankovich, V., Schoenberg, M., Izadi, S., Funkhouser, T., Fanello, S., 2018. Activestereonet: End-to-end self-supervised learning for active stereo systems .

[35] Zhang, Z., 1999. Flexible camera calibration by viewing a plane from unknown orientations, in: Computer Vision, 1999. The Proceedings of the Seventh IEEE International Conference on.

[36] Zhong, Y., Dai, Y., Li, H., Zhong, Y., Dai, Y., Li, H., 2017. Self-supervised learning for stereo matching with self-improving ability .